\documentclass[letterpaper, 10 pt, conference]{ieeeconf}  
\usepackage{graphicx}
\graphicspath{ {images/} }

\IEEEoverridecommandlockouts                              
\title{\LARGE \bf
Simultaneous Localization And Mapping with depth Prediction using Capsule Networks for UAVs
}

\author{Sunil Prakash$^{1}$ and Gaelan Gu$^{2}$
\thanks{$^{1}$S. Prakash is with Institute of Systems Science, National University of Singapore, Singapore}
\thanks{$^{2}$G. Gu is with Institute of Systems Science, National University of Singapore, Singapore}
}

\begin{document}

\maketitle
\thispagestyle{empty}
\pagestyle{empty}

\begin{abstract}

In this paper, we propose an novel implementation of a simultaneous localization and mapping (SLAM) system based on a monocular camera from an unmanned aerial vehicle (UAV) using Depth prediction performed with Capsule Networks (CapsNet), which possess improvements over the drawbacks of the more widely-used Convolutional Neural Networks (CNN). An Extended Kalman Filter will assist in estimating the position of the UAV so that we are able to update the belief for the environment. Results will be evaluated on a benchmark dataset to portray the accuracy of our intended approach.

\end{abstract}

\section{INTRODUCTION}

Simultaneous localization and mapping (SLAM) has a significant role to play in helping autonomous robots to navigate their way around an uncertain environment and this has many widespread implications in various industries. For instance, drones can be programmed to find their way in a logistics warehouse without prior knowledge of the space, in order to retrieve information of a particular package. The fact that this can be performed using a single camera expands the realm of possibilities SLAM can have with a low-cost approach that is quickly gaining traction.

In order to implement SLAM, depth maps will need to be created based on images recorded on the camera so that accurate tracking of the unmanned aerial vehicle (UAV) can be achieved. Currently deep Convolutional Neural Networks (CNNs) are popularly used in depth prediction with monocular camera systems due to the high accuracy which can be achieved in the image recognition space. Lower-level neurons send information about the objects in an image to higher-level neuron, where they perform further convolutions to identify these objects using features present in the image. However CNNs have a severe limitation whereby they are only able to register if the objects exist in the image; they do not encode the orientation and position of objects. This issue had been mentioned by Geoffrey Hinton, one of the founding fathers of deep learning, and he has proposed an alternative deep neural network which can address this drawback.

Capsule Networks (CapsNets) consist of capsules, or a group of neurons, to identify patterns in an image at the lower-level layers. This information comes in the form of high-dimensional vectors containing probabilities of the orientation and positions of patterns residing in the image, which is then received by higher-level capsules. The higher-level capsules process this information from several lower-level capsules and subsequently outputs a prediction. As the CapsNet has achieved a greater accuracy than the CNN on widely-available datasets like MNIST, we hope to attain a similar performance by applying the CapsNet with our SLAM method.

\section{RELATED WORK}

Previous work include those of Chatterjee, A. \cite{c1} for using a Neuro-Fuzzy EKF-based approach for SLAM problems, and Ventura, J. \cite{c3} for performing localization using mobile phone. Most of these projects were done mostly with a keyframe-based approach, where if a good set of points in the image stream is detected to be tracked in the next frame, those points are designated as \textit{keypoints}. A good keyframe algorithm used in earlier works is called the Harris Corner Detector \cite{c8}. The basic idea is to come up with an image region which could be used for tracking if there are shifts in all directions. Mathematically this could be formulated by recording any change in approaches when shifting the window by (x,y) large for every (x,y) on the unit circle. This change is a weighted sum of square differences and denoted by S(x,y). When used with Faussian kernel with fixed variance ${\sigma ^2}$, it could be written as:

\begin{equation}
S(x,y) = \sum _{u,v} w(u,v) (I(u_x, v+y) - I (u,v))^2
\end{equation}

Further, a few more algorithms which are also good in detecting keypoints like FAST Corner detector presented by Roston and Drummond in 2006 \cite{c8} and Laplacian of Gaussian (LoG) blob detector \cite{c8} which finds blobs in image instead of corners identified using Laplacian on the smoothed image.

Additionally, CNN based SLAM \cite{c2} can be used for identifying the depth predictions and feeding information to the keyframe initializer and for semantic label fusion, but efficiently most of the work is done using the depth or stereo camera. In particular, the predicted depth map is used as input for Kellar's Point-Based Fusion (RGB-D SLAM) algorithm, but it lacks sharp details mostly due to blurring artifacts.

\begin{figure*}
\includegraphics[width=\textwidth]{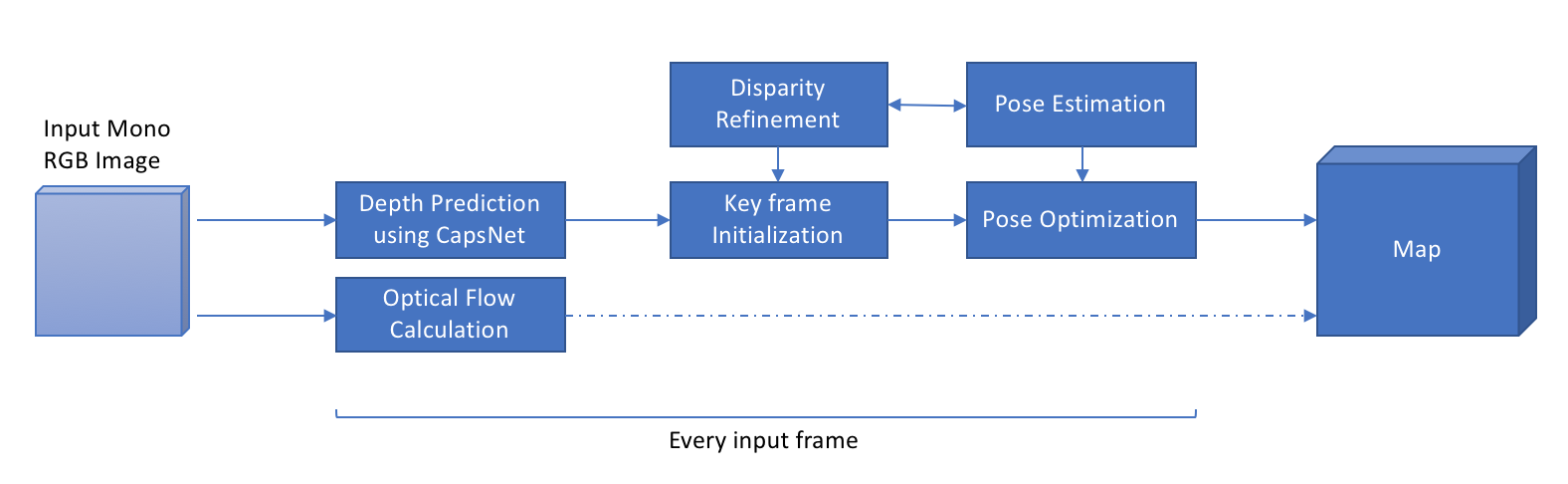}
\caption{Proposed SLAM Overview}
\label{fig:fig1}
\end{figure*}

\section{PROPOSED UAV SLAM}
In this section, we explain the overview of proposed framework for Map construction by reconstructing disparity map from the single image and it is fused together with keyframe detected and optical flow calculation to create the Global map of the environment as seen in the Figure \ref{fig:fig1}. Interestingly we explain how monocular image could be used to find the depth of the image.

Using the CNN-based approach is slow and lots of image pre-processing is required due to a lack of information processing with CNN. To maintain a high accuracy of the depth, we propose a depth map via CapsNet on the entire frame. Moreover an uncertainty map is constructed by measuring the pixel-wise confidence of each disparity produced.

\subsection{Camera Pose Estimation}
The camera pose estimation is based on the keyframe approach \cite{c1}. In particular, all potentially visible landmarks are projected into the image and expected camera position ${C_0}$, and for each landmark, a warped template is generated from the depth map ${D_k}$ for each keyframe ${k_{1},...,k_{n} \in K}$. At each frame ${t}$ , we estimate the camera pose ${C^{ki}_{t}}$ , i.e the transformation among nearest point ${k_i}$ and frame ${t}$, formed by square rotation matrix and a 3D translation vector. The expected camera pose then could be described as:

\begin{equation}
E(C^{ki}_{t}) = \sum \rho (\frac{r(u,C^{ki}_{t} )}{\sigma(r(u,C^{ki}_{t}))}) 
\end{equation}

where ${\rho}$ is the Huber norm and ${\sigma}$ is a function measuring the residual uncertainty. ${r}$ is the optical residual defined by unmapped pixels from 2D points to 3D coordinates.

\subsection{Capsule Network based Depth Prediction}

As described by Hinton in the paper defining dynamic routing between capsules \cite{c9} within the capsule network.

\begin{figure}
\includegraphics[scale=0.5]{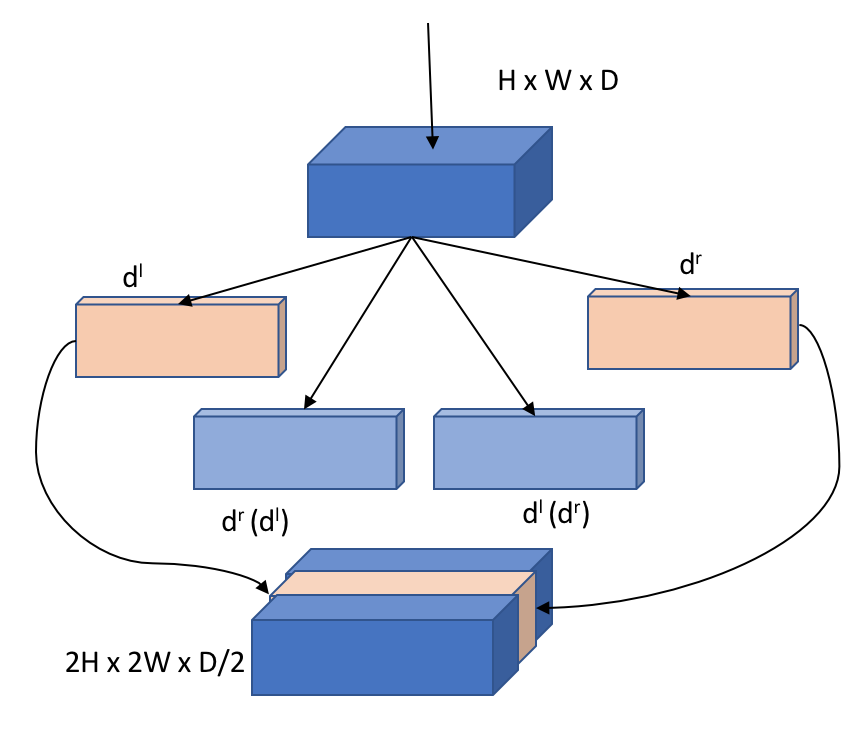}
\label{fig:fig2}
\caption{Depth Estimation}
\end{figure}

Capsules are nested set of neural network layers. In a normal neural network, in order to make it a deep network, it is required to add more and more layers while sometimes skipping connections between the layers. On the other hand for a CapsNet, more layers are added inside a single layer. The state of the neurons inside the image captures the property of one region or entity inside the image. The vector is then shared to all possible parents of the neural network and a prediction vector is calculated by multiplying its own weight and layers' weight matrix. The parent which calculates the largest prediction vector product, increases the capsule bond while the others decrease their bonds. It is an interesting way of capturing most of the attributes of the section of focus in the image, free from sparsity and scaling issues as seen with CNNs.

We were inspired with the approach described by Clement Godard \cite{c10} where depth of the image was calculated using left-right consistency. Similarly, we proposed a novel approach of treating depth estimation problem as an image reconstruction problem, so we can calculate the disparity field without requiring the ground truth depth. For this we trained our capsule network with the pairs of images generated from the stereo camera where left camera image is used as input to generate an image which could be similar as right camera image.

A single image ${I}$ is taken at the test time and the objective is to learn a function ${g}$ which could predict the per-pixel depth, ${\hat{d} = g(I)}$. And while training, we feed in two images ${I^l}$ and ${I^r}$, for left and right images of the stereo camera images captured at the same moment of time. We reconstruct the image ${I^l(d^r)}$ as ${\hat{I^r}}$ and vice versa. Given with the baseline distance ${z}$ between the camera with focal length ${f}$ , we could recover the depth ${\hat{d}}$ from the predicted disparity from the neural network as ${\hat{d} = zf/d}$
    
The loss function for disparity map prediction used by CapsNet as ${L_s}$ could defined as:

\begin{equation}
L_s = \alpha_{ap}(L^l_{ap} + L^r_{ap}) + \alpha_{ds}(L^l_{ds} + L^r_{ds}) + \alpha_{lr}(L^l_{lr} + L^r_{lr}) + \zeta
\end{equation}

where ${L_{ap}}$ is the loss from reconstructing the image, ${L_{ds}}$ is loss from smooth disparity adjustment , ${L_{lr}}$ is loss from predicted disparity and ${\zeta}$ is the regularizing constant.

\begin{figure}[h]
\includegraphics[scale=0.23]{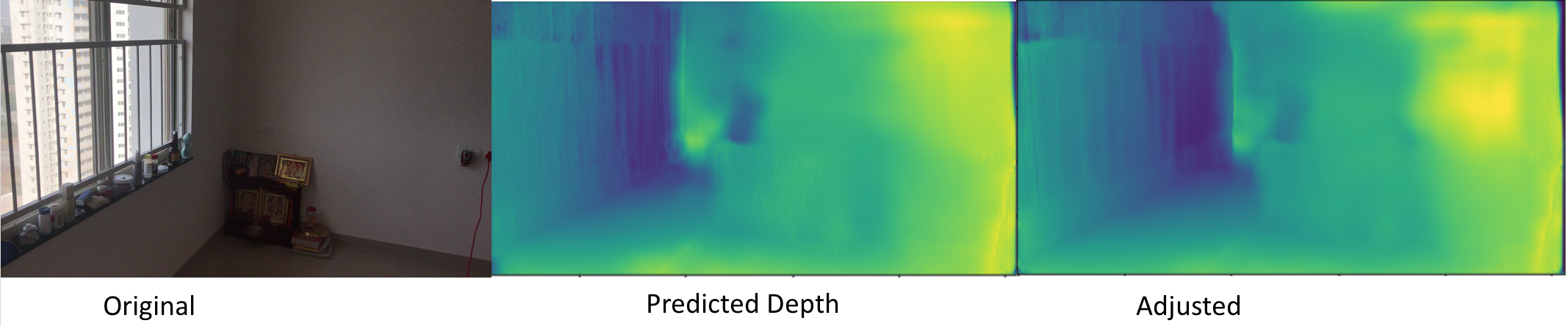}
\includegraphics[scale=0.23]{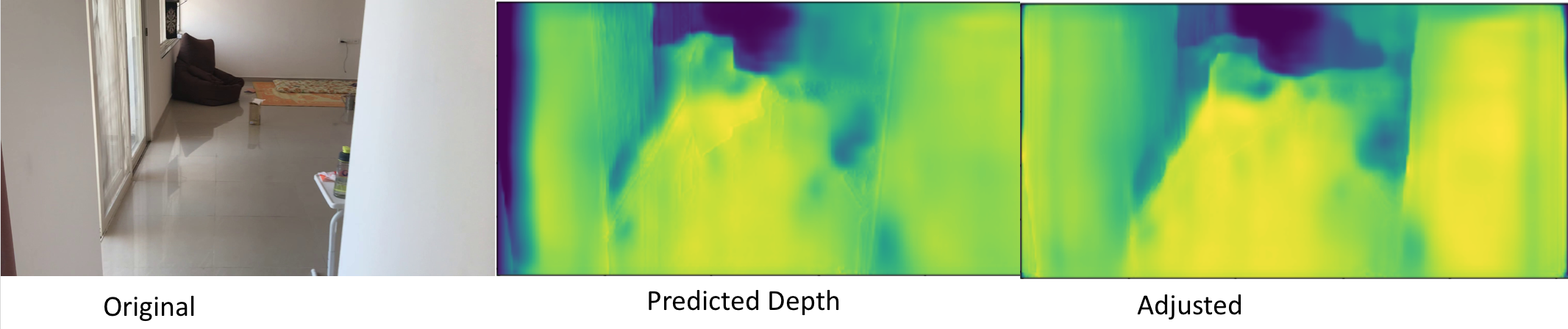}
\caption{Depth Image Generation}
\label{fig:fig3}
\end{figure}

The result of the generated image could be seen in Figure \ref{fig:fig3} where depth could be interpreted as a darker shade of color. The darker the shade, the more depth there is in that image section.

\subsection{Extended Kalman Filter}
The Kalman Filter is a popular method for fusing noise measurement of a dynamic system to get a good estimate of current state. It assumes the system to be linear and any dynamic model could be predicted efficiently if said system is linear. But since most real-world problems are non-linear systems, the Extended Kalman Filter (EKF) seems to be more appropriate for handling such problems.

\section{SYSTEM ARCHITECTURE}

In this section we discuss about the overall system architecture and component integration with implementation details.

\subsection{Hardware}

We used a Parrot AR Quadcopter Drone, equipped with a Robotic Operating System (ROS). The drone has also been fitted with a forward-facing camera capable of 18fps, 640 x 480 pixel footage, downward-facing camera capable of 60fps, 176 x 144 pixel footage, ultrasound sensor, 3-axis accelerometer and 2-axis gyroscope.

For training the CapsNet, we used an Amazon EC2 P3 instance with 32GB RAM, 8GB GPU memory and 1TB disk space.

\subsection{Datasets}

Data from the Technical University of Munich (TUM) was used as a benchmark for our SLAM method. The dataset prepared by the Computer Vision Group in the Faculty of Informatics consists of 50 video sequences with a combined duration of more than 100 minutes, recorded across various real-world environments - both narrow indoor corridors and wide outdoor areas. These were captured using a monocular camera situated on a drone and photometric calibration was also performed on the images to provide additional information such as exposure time and camera response function. 

We have also prepared our own data by capturing sequences of an apartment interior with a Our drone. For the intention of our work, only the forward-facing camera was used to record short sequences of a few minutes within a controlled indoor environment. The captured sequences were then used to determine if the network trained on the TUM dataset generalizes well on a new dataset.

\subsection{UAV System Integration}

An unsupervised learning approach with the CapsNet will produce a depth or disparity map, based on the images provided. This depth map will subsequently be employed for recording and tracking purposes, as the UAV traverses across the environment. Optical flow can also be calculated to measure the pattern of apparent motion of objects in a visual scene, due to the motion of the camera. This will be applied onto the disparity map to detect the direction of the camera. Following which, an EKF algorithm can estimate the position of the UAV. We will finally produce a mapped reproduction of the environment.

\begin{figure}[h]
\includegraphics[scale=0.5]{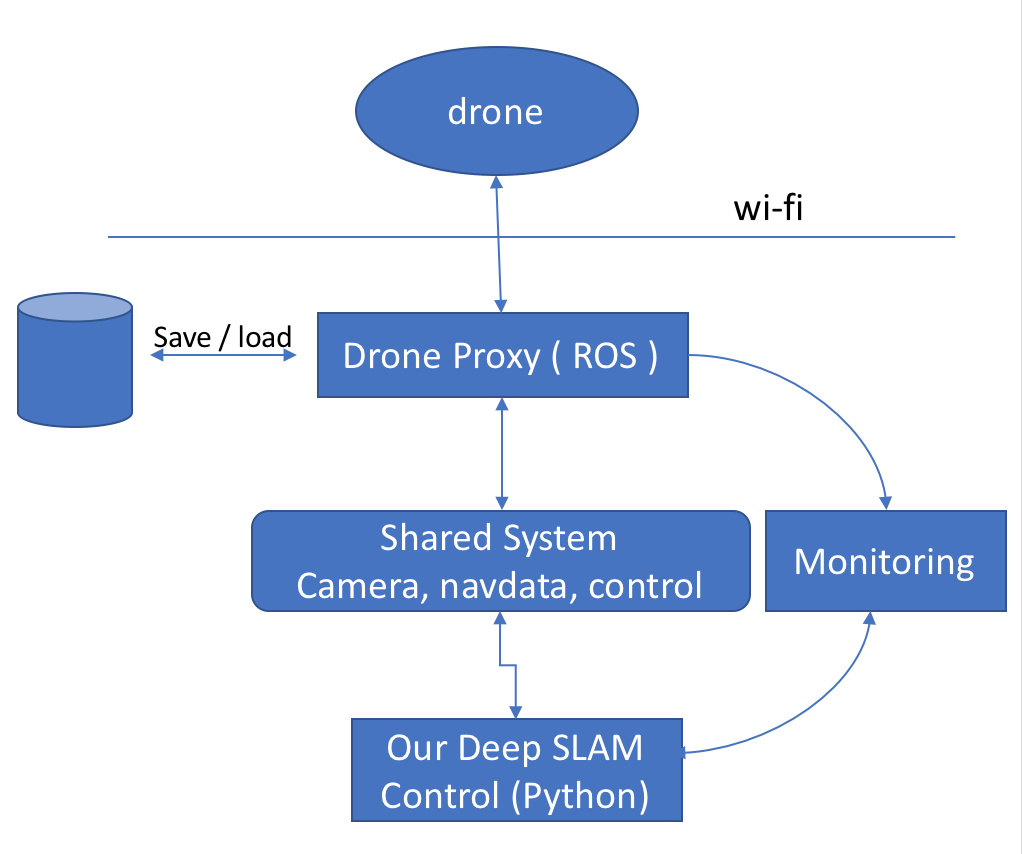}
\caption{UAV System Architecture}
\label{fig:fig4}
\end{figure}

\section{EXPERIMENT}
In this section we discuss about various experiments done to achieve the state of the art results using various algorithms used.

We compared our results against the publicly available result of ORB-SLAM \cite{c10} and CNN-SLAM \cite{c2}, both are used for feature-based methods where one relies on image edges based keyframes and the other on features generated using the CNN network and image segmentation. We have also compared our result with Direct Sparse Odometry SLAM which is a version of dense SLAM that is more computationally heavy.

\begin{table}[ht]
\begin{tabular}{|p{1.5cm}||p{1.5cm}|p{1.5cm}|p{2cm}|}
 \hline
 \multicolumn{4}{|c|}{Perc. Correct Depth} \\
 \hline
 Dataset  & ORB-SLAM & CNN-SLAM & Our Method\\
 \hline
 TUM/seq1 & 0.031 & 12.477 & 12.369 \\
 TUM/seq2 & 0.059 & 24.077 & 25.784 \\
 TUM/seq3 & 0.027 & 27.396 & 29.980 \\
 Our data & 0.019 & 23.059 & 28.590 \\
 \hline

\end{tabular}
\label{table:tab1}
\caption{Comparison Matrix of Percentage of Correctly Estimated Depth}
\end{table}

The results in Table \ref{table:tab1} show that our method achieves a higher level of accuracy in most cases, with the exception of the TUM sequence 1. In general, we can say that the CapsNet does a better job at predicting depth than the other methods and we would attribute this result to the above-mentioned advantages of using the network compared to the CNN for instance.

\begin{figure}
\includegraphics[scale=0.5]{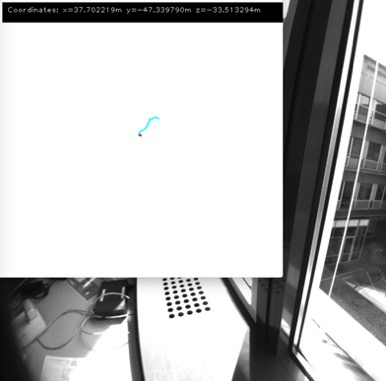}
\caption{Creating the SLAM Method}
\label{fig:fig7}
\end{figure}

\begin{figure}[t]
\includegraphics[scale=0.5]{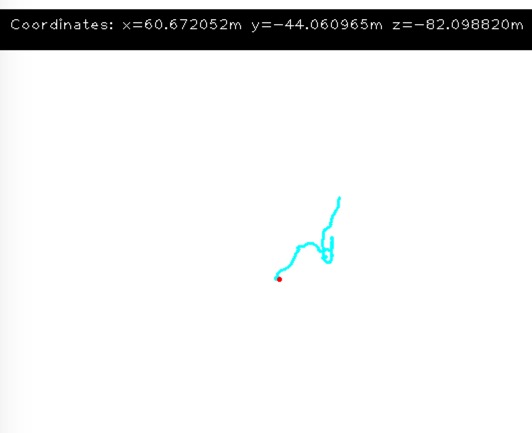}
\caption{SLAM Mapping}
\label{fig:fig8}
\end{figure}

\section{CONCLUSIONS AND FUTURE WORK}

Through our work, we have shown how the use of CapsNets over CNNs in depth estimation for SLAM problems has much potential in programming autonomous UAVs. Even when faced with limitations related to the single-camera system, depth maps obtained from the network can be reliable enough for further application.

Future improvements could include the fusion of additional sensors in order to improve the overall accuracy. For example, the accelerometer can pinpoint the pose and direction of the UAV and the gyroscope could be used to orientation of the drone. The second single camera present on the UAV can also be used to generate another set of depth maps from a different perspective. The data gathered from the other sensors will have to be processed differently from that of the images.

\addtolength{\textheight}{-12cm}   




\section*{ACKNOWLEDGMENT}

Thanks to Tian Jing for his invaluable advice and guidance throughout the project. This project was part of a final-year research elective and is supported by the Institute of Systems Science, National University of Singapore.



\begin{thebibliography}{99}

\bibitem{c1} Chatterjee, A., Matsuno, F. (2007). A Neuro-Fuzzy Assisted Extended Kalman Filter-Based Approach for Simultaneous Localization and Mapping (SLAM) Problems. IEEE Transactions on Fuzzy Systems, 15(5), 984-997. https://doi.org/10.1109/TFUZZ.2007.894972.
\bibitem{c2} Tateno, K., Tombari, F., Laina, I., Navab, N. (2017). CNN-SLAM: Real-time dense monocular SLAM with learned depth prediction. 2017 IEEE Conference on Computer Vision and Pattern Recognition (CVPR), Honolulu, HI, 6565-6574. https://doi.org/CVPR.2017.695.
\bibitem{c3} Ventura, J., Arth, C., Reitmayr, G., Schmalstieg, D. (2014). Global localization from monocular SLAM on a mobile phone. IEEE Trans Vis Comput Graph, 20(4), 531-539. https://doi.org/10.1109/TVCG.2014.27.
\bibitem{c4} Mur-Atal, R., Montiel, J. M. M., Tardós, J. D. (2015). ORB-SLAM: A Versatile and Accurate Monocular SLAM System. IEEE Transactions on Robotics, 31(5), 1147-1163. https://doi.org/10.1109/TRO.2015.2463671.
\bibitem{c5} Williams, B., Klein, G., Reid, I. (2007). Real-time SLAM relocalisation. IEEE 11th International Conference on Computer Vision, 1-8. https://doi.org/10.1109/ICCV.2007.4409115.
\bibitem{c6} Huang, T. W., Hsu, C. C., Wang, W. Y., Baltes, J. (2017). ROSLAM—A Faster Algorithm for Simultaneous Localization and Mapping (SLAM). Robot Intelligence Technology and Applications 4. Advances in Intelligent Systems and Computing, 447. Springer, Cham.
\bibitem{c7} Kung, D. W., Hsu, C. C., Wang, W. Y., Baltes, J. (2017). Adaptive Computation Algorithm for Simultaneous Localization and Mapping (SLAM). Robot Intelligence Technology and Applications 4. Advances in Intelligent Systems and Computing, 447. Springer, Cham.
\bibitem{c8} Shuvalova, L., Petrov, A., Khithov, V., Tishchenko, I. (2017). Interactive Markerless Augmented Reality System Based on Visual SLAM Algorithm. Robot Intelligence Technology and Applications 4. Advances in Intelligent Systems and Computing, 447. Springer, Cham.
\bibitem{c9} Sara Sabour, Nicholas Frosst, Geoffrey E Hinton. (2017) Dynamic Routing Between Capsules \textit{	arXiv:1710.09829}
\bibitem{c10} Clement Godard, Oisin Mac Aodha, Gabriel J. Brostow. (2017) Unsupervised Monocular Depth Estimation with Left-Right Consistency \textit{	arXiv:1609.03677}
\bibitem{c11} Georg Klein, David Murray. Parallel Tracking and Mapping for Small AR Workspaces (2007)

\end{thebibliography}
\end{document}